\documentclass[10pt,journal,compsoc]{IEEEtran}

\usepackage{cite}
\usepackage{amsmath,amssymb,amsfonts}
\usepackage{graphicx}
\usepackage{epsfig}
\usepackage{epstopdf}
\usepackage{textcomp}
\usepackage{mathtools}
\usepackage{xcolor}
\usepackage{tikz}
\usetikzlibrary{calc}
\usepackage{soul}
\soulregister\cite7
\soulregister\ref7
\usepackage{wrapfig}
\usepackage{multirow}

\usepackage[utf8]{inputenc} 
\usepackage[T1]{fontenc}    
\usepackage{hyperref}       
\usepackage{url}            
\usepackage{booktabs}       
\usepackage{amsfonts}       
\usepackage{nicefrac}       
\usepackage{microtype}      
\usepackage{threeparttable}
\usepackage{multirow}
\usepackage{graphicx}
\usepackage[square,numbers]{natbib}
\usepackage{amsmath}
\usepackage{color}
\usepackage{subfigure}
\usepackage{makecell}
\usepackage{multicol}
\usepackage{listings}
\usepackage{wrapfig}
\usepackage[linesnumbered, ruled]{algorithm2e} 
\usepackage{algorithmic}

\newcommand{\name}{\textsc{FenceBox}\xspace}

\begin{document}

\title{\name: A Platform for Defeating Adversarial Examples with Data Augmentation Techniques}

\author{Han~Qiu,~\IEEEmembership{Member,~IEEE,}
        Yi~Zeng,
        Tianwei~Zhang,
        Yong~Jiang,
        and~Meikang~Qiu,~\IEEEmembership{Senior~Member,~IEEE}
        
\IEEEcompsocitemizethanks{
\IEEEcompsocthanksitem H. Qiu is with Telecom Paris, Institut Polytechnique de Paris, Palaiseau, France, 91120. Email: han.qiu@telecom-paris.fr.
\IEEEcompsocthanksitem Y. Zeng is with University of California San Diego, CA, USA, 92122. Email: y4zeng@eng.ucsd.edu.
\IEEEcompsocthanksitem T. Zhang is with Nanyang Technological University, Singapore, 639798. Email: tianwei.zhang@ntu.edu.sg.
\IEEEcompsocthanksitem Y. Jiang is with Tsinghua Shenzhen International Graduate School, Tsinghua University, China, 518055. Email: jiangy@sz.tsinghua.edu.cn.
\IEEEcompsocthanksitem M. Qiu (corresponding author) is with the Texas A\&M University, Texas, USA, 77843. Email: meikang.qiu@tamuc.edu
}
\thanks{Manuscript received xxx.}
}


\IEEEtitleabstractindextext{%
\begin{abstract}
It is extensively studied that Deep Neural Networks (DNNs) are vulnerable to Adversarial Examples (AEs). 
With more and more advanced adversarial attack methods have been developed, a quantity of corresponding defense solutions were designed to enhance the robustness of DNN models. It has become a popularity to leverage data augmentation techniques to preprocess input samples before inference to remove adversarial perturbations. By obfuscating the gradients of DNN models, these approaches can defeat a considerable number of conventional attacks. Unfortunately, advanced gradient-based attack techniques (e.g., BPDA and EOT) were introduced to invalidate these preprocessing effects. 

In this paper, we present \name, a comprehensive framework to defeat various kinds of adversarial attacks. \name is equipped with 15 data augmentation methods from three different categories. We comprehensively evaluated that these methods can effectively mitigate various adversarial attacks. \name also provides APIs for users to easily deploy the defense over their models in different modes: they can either select an arbitrary preprocessing method, or a combination of functions for a better robustness guarantee, even under advanced adversarial attacks. We open-source \name, and expect it can be used as a standard toolkit to facilitate the research of adversarial attacks and defenses. 

\end{abstract}

\begin{IEEEkeywords}
Deep Learning, Adversarial Examples, Security, Robustness, Data Augmentation.
\end{IEEEkeywords}}

\maketitle

\IEEEdisplaynontitleabstractindextext

\IEEEpeerreviewmaketitle

\section{Introduction}
\label{sec:introduction}

Modern software systems and applications are becoming more intelligent and omnipotent, due to the revolutionary development of Deep Learning (DL) technology. A large quantity of DL models, algorithms, and techniques have been proposed to solve more complex tasks and meet people's increasing demands on Artificial Intelligence. These DL-based inventions and products significantly improve the quality of our life and promote social progress. 

Unfortunately, Deep Learning also brings new security challenges in addition to the benefits. One of the most severe security threats against DL systems is the Adversarial Example (AE) \cite{szegedy2013intriguing}: with imperceptible and human unnoticeable modifications to the input, a DL model can be fooled to give wrong prediction results. This threat is caused by the distinctions of understanding and interpretation between humans and machines: although DL models have satisfactory performance in terms of automation, speed, and possibly accuracy, they can make simple mistakes that humans will never do. 

The severity of AEs exhibits in two directions. First, as DL techniques have been widely commercialized in various products, AEs can threaten these DL applications and bring catastrophic consequences to our life. Past works have demonstrated the practicality of such attacks in many security- and safety-critical domains, e.g., autonomous driving \cite{cao2019adversarial,zhao2019seeing,eykholt2018robust}, home automation \cite{carlini2016hidden,yuan2018commandersong,zhang2017dolphinattack}, etc. Second, different from traditional intrusion attacks, AEs can be conducted by just manipulating the input (e.g., sensory data) slightly. Therefore, most conventional intrusion detection and prevention mechanisms such as access control, cryptographic encryption, and protocols are ineffective to counteract this type of threat. It is in urgent need of new defense solutions to thwart AEs and protect DL systems.  

Over the past years, a great number of works focused on the defenses against AEs generated from different attack techniques. These solutions can be roughly classified into the following categories: (1) adversarial training~\cite{kurakin2016adversarial, huang2015learning, shaham2018understanding}: AEs are used with normal examples together to train DNN models to recognize and correct malicious samples. (2) Training assisted models to distinguish malicious samples from benign ones, such as Magnet~\cite{meng2017magnet} and Generative Adversarial Trainer~\cite{lee2017generative}. (3) Designing AE-aware network architecture or loss function, e.g., Deep Contractive Networks~\cite{gu2014towards}, Input Gradient Regularization~\cite{ross2018improving}, Defensive Distillation~\cite{papernot2016distillation}, etc. However, those solutions require significant modifications to the original DNN model, which can bring inconvenience and increase the computation cost, especially for large-scale DNN models. Besides, these methods lack generality to cover various types of attacks. They ``explicitly set out to be robust against one specific threat model'' \cite{carlini2019evaluating}.

In contrast, data augmentation based solutions can overcome the above limitations. These solutions \cite{guo2018countering,prakash2018deflecting,xie2018mitigating,das2018shield} introduce a preprocessing function over the input samples before feeding them into the DNN models. Such preprocessing operation can either remove the effects of adversarial perturbations on the inference or make it infeasible for the adversary to generate AEs adaptively, even he knows every detail of the operation. A good preprocessing function should be designed as (1) general-purpose to cover different types of models and defeat different types of attacks; (2) lightweight with negligible computation cost to the inference pipeline; (3) usability-preserving without decreasing the model's prediction accuracy. 

In this paper, we design \name, a first-of-its-kind platform to thwart adversarial examples using preprocessing techniques. Some past works have developed platforms and toolkits for evaluating and analyzing adversarial attacks and defenses \cite{ling2019deepsec,papernot2016technical,nicolae2018adversarial}. Those works collected both attacks and different types of defense solutions. In contrast, we mainly focus on the preprocessing-based solutions and identify a lot of effective solutions that were never used for adversarial defense previously.

Specifically, \name incorporates 15 data augmentation functions which can satisfy the above requirements. They are classified into three categories based on the characteristics of operations: image distortion, compression, and noise injection. We present a systematic evaluation of the effectiveness of different preprocessing solutions against various adversarial attacks. We consider both standard attacks (FGSM, IFGSM, C\&W, and LBFGS) and advanced attacks (PGD and BPDA). For each preprocessing solution, we measure its impact on the clean samples, as well as the success rate of each attack with different configurations. Such comprehensive evaluation reveals the relative effectiveness of each operation and technique.


This platform can be readily used to protect critical DL applications and systems. It can be applied to different scenarios. A cloud provider can adopt this platform to set up a ``preprocessing-as-a-service'' to purify the images for end users. In edge computing, the administrator can implement this platform between the sensors (e.g., cameras) and the DL inference engine, to remove potential malicious perturbations. This platform is flexible and user-friendly, with different usage modes. Users can select the optimal preprocessing technique based on their threat models and demands. They can also choose an ensemble of multiple techniques to enhance defense strength. Our evaluations indicate these modes can significantly increase the difficulty and cost of AE generations. We open source \name online to better promote this research direction\footnote{\url{https://github.com/YiZeng623/FenceBox}}. 
We hope this can benefit future research in adversarial attacks, defenses, and analysis, and also help practitioners to secure their DL applications. We also encourage the public to contribute to the further development of this platform. 

The rest of the paper is organized as follows: Section \ref{sec:bg} gives the background of adversarial attacks and defenses. Section \ref{sec:design} discusses the threat model and defense requirements. Section \ref{sec:overview} presents detailed descriptions and categorization of the data augmentation methods in our consideration. Comprehensive evaluations of these techniques against various adversarial attacks are given in Section \ref{sec:evaluation}. Section \ref{sec:case} provides a case study to illustrate the usage of \name. We conclude in Section \ref{sec:conclusion}.

\section{Research Background}
\label{sec:bg}

In this section, we first briefly introduce the definition and development of the adversarial examples against DL models. 
Then, the development and categorization of the defense techniques is reviewed and discussed. 
The existing state-of-the-art data augmentation based defense methods are introduced in the end.


\subsection{Adversarial Attacks against DNN Models}

In an adversarial attack, the adversary tries to add human-unnoticeable perturbations on the original input to fool a DNN classifier. 
Formally, the target DNN model is a mapping function $F(\cdot)$. 
Given a clean input sample $x$, the corresponding AE is denoted as $\widetilde{x}=x+\delta$ where $\delta$ is the adversarial perturbation. 
Then AE generation can be formulated as the optimization problem in Eq. (\ref{eq:AE}a) as a targeted attack where $l'\neq F(x)$ is the desired label set by the attacker or in Eq. (\ref{eq:AE}b) as an untargeted attack. 

\begin{subequations}
\label{eq:AE}
  \begin{align}
& min  \lVert\delta\rVert, s.t. \: F(\widetilde{x})=l' \\
& min  \lVert\delta\rVert, s.t. \: F(\widetilde{x})\neq F(x)
  \end{align}
\end{subequations}

Generally, there are two attack scenarios~\cite{carlini2019evaluating}, determined by the adversary's knowledge about the target system.
(1) \emph{White-box scenario}: the adversary knows all details about the target DNN model including the architecture and all the parameters. 
He is also aware of the defense mechanism and the corresponding parameters. 
(2) \emph{Black-box scenario}: the adversary does not have any knowledge about the target DNN system. 
In addition to these two scenarios, there are also some works~\cite{prakash2018deflecting} considering the \emph{gray-box scenario} where the adversary knows all details about the model but not the defense mechanism. 
It is not quite realistic and reasonable to assume the secrecy of adopted defenses, as ``this widely held principle is known in the field of security as Kerckhoffs' principle.''~\cite{carlini2019evaluating}. 
So we exclude this gray-box scenario in this paper. 

Various methods have been proposed to generate AEs, and they can be classified into three main categories according to the development phases of the attack techniques. 

\noindent\textbf{Basic gradient-based approaches}: the adversary generates AEs by calculating the model gradients with pre-set constraints of perturbation scales. 
For instance, Fast Gradient Sign Method (FGSM)~\cite{goodfellow2014explaining} calculates the perturbations based on the sign of the gradient of the loss function with respect to the input sample. 
Later on, some improved methods are given based on FGSM such as I-FGSM~\cite{kurakin2016adversarial} and MI-FGSM~\cite{dong2017discovering}. They all aim at iteratively calculating the adversarial perturbations based on FGSM with a small step or with momentum. 

\noindent\textbf{Optimized gradient-based approaches}: the adversary adopts optimization algorithms~\cite{carlini2017towards} to find optimal perturbations by considering the gradients of the output prediction results with respect to input samples. 
This kind of optimized gradient-based attacks is proved to be very effective in a white-box scenario since the attackers can optimize the AE generation according to the defense methods. 
For instance, by carefully setting the optimization functions, C\&W~\cite{carlini2017towards} can find the input features that make the most significant changes to the final softmax output to mislead the DNN models. 

\noindent\textbf{Approximated gradient-based approaches}: the adversary can approximate the target DNN model's gradients to generate the AEs. 
Such approaches were proposed to defeat some defense methods which obfuscate the gradients via random or  non-differentiable operations. For instance, Backward Propagation Differentiable Approximation (BPDA)~\cite{athalye2018obfuscated} approximates the shattered gradients by approximating the non-differentiable and non-invertible layers in the DNN inference process. 
Expectation of Transformation (EOT)~\cite{athalye2018obfuscated} approximates the gradients by calculating the average of the random gradients generated by the layers with random gradients over a quantity of inference sessions.

\subsection{Defense Techniques against Adversarial Attacks}

In the past several years, various defense solutions were also proposed to mitigate the corresponding attacks \cite{qiu2020mitigating}. 
We briefly review the development of these defense approaches as three categories.

\noindent\textbf{Detecting AEs.}
A popular direction is to compare the internal model behaviors (e.g., activated neurons) to distinguish AEs from normal samples. However, such approaches inevitably introduce false positives, especially for those benign samples with low prediction confidence. 
How to distinguish these normal yet low-confidence samples from real AEs is still unknown. 
Besides, once an AE is identified, how to automatically repair the samples and recover the correct results is still not solved. 

\noindent\textbf{Reinforcing DNN models.}
The idea of these approaches is to enhance the robustness of DNN models while maintaining high accuracy on normal samples. As shown in Eq.~\ref{eq:model}, $F'$ denotes the mapping function of the reinforced DNN model. 
The strategy attempts to design a $F'$ to classify each AE $\widetilde{x}$ correctly.

\begin{equation}
    \max_{F'} P\; (F'(\widetilde{x})=F'(x))
    \label{eq:model}
\end{equation}

One typical example of this strategy is network distillation~\cite{papernot2016distillation}, which was broken by generating the corresponding AEs~\cite{carlini2017towards} to defeat the reinforced DNN model $F'$ once the attacker has the knowledge of this new $F'$. Another example is adversarial training~\cite{goodfellow2014explaining,madry2017towards}. This is proved to be very effective against many known attacks~\cite{raff2019barrage}, and could be the most encouraging and desirable path toward defending against adversarial attacks in the long term~\cite{raff2019barrage}. However, such a method is not yet practically usable for large datasets and complex DNN models. 

The third defense category is input preprocessing, as described below. 

\subsection{Input Preprocessing as a Defense}
The goal of this strategy is to apply a transformation function on the input samples before the inference. This function is expected to reduce the impact of carefully-crafted perturbations on the model prediction~\cite{qiu2020review}. Formally, we aim to identify a function $g$, which can maximize the probability of classifying the adversarial examples $\widetilde{x}$ to the right class as the original sample $x$, as shown in Eq.~\ref{eq:transform}:

\begin{equation}
    \max_{g} P\; (F(g(\widetilde{x}))=F(g(x)))
    \label{eq:transform}
\end{equation}

Commonly, a qualified function should be non-differentiable and non-invertible in order to thwart the adversary from obtaining the gradients of the target model~\cite{prakash2018deflecting}. Different preprocessing functions have been utilized to achieve this goal. The first approach is image distortion. For instance, \cite{xie2018mitigating} proposed to randomize the scale of image content by changing the pixel locations and dropping a small portion of pixel values. The second approach is image compression. 
\cite{liu2019feature} quantized the pixel values or the corresponding frequency coefficients into a set of fixed values to remove the small perturbations added by attackers.
The third approach is to inject artificial noise on the input images to eliminate the effects of the adversarial perturbations. \cite{prakash2018deflecting} proposed to first inject random noise on AEs, and then use denoise filters to increase the classification accuracy. 

These gradient obfuscation approaches are effective against the basic and optimized gradient-based attacks. They fail to defeat approximated gradient-based techniques. A recent work~\cite{raff2019barrage} proposed to randomly select an ensemble of these simple preprocessing functions to mitigate those advanced attacks. In this paper, we aim to perform a large-scale evaluation systematically and identify more powerful preprocessing functions and integration.




\section{Preliminary}
\label{sec:design}

In this section, we introduce the threat model and defense requirements considered in this paper. 


\subsection{Threat Model}

First, we only evaluate the targeted attacks~\cite{carlini2017towards}: attacks succeed only when the DNN model predicts the input as the specific label desired by the adversary. The untargeted attack in which the adversary tries to mislead the DNN models to an arbitrary label different from the correct one can be mitigated in a similar way. We do not discuss this case in this paper. 

Second, we consider our defense in a white-box scenario where the adversary has full knowledge of the DNN model, including the network architecture, exact values of model parameters, and all hyper-parameters. 
We also assume that the adversary has full knowledge of the proposed defense, including the algorithms and parameters. 
For the defenses involving randomization operations, we assume the random numbers generated in real-time are perfect with a large entropy such that the adversary cannot obtain or guess the correct values. 
It is worth noting that this white-box scenario represents the strongest adversaries. 
Under this setting, a big number of existing state-of-the-art defenses can be invalidated as shown in~\cite{tramer2020adaptive}. 
This significantly increases the difficulty of defense designs. 

Third, we consider the attacker's capabilities as follows. He is outside of the DNN classification system and cannot compromise the inference computation or the DNN model parameters (e.g., via fault injection to cause bit-flips \cite{rakin2019bit} or backdoor attacks \cite{gao2019strip}). He can only manipulate the input data with imperceptible perturbations. 
He can directly modify the input image pixel values within a certain range. We use $l_{\infty}$ and $l_{2}$ distortion metrics to measure the scale of added perturbations: we only allow the generated AEs to have either a maximum $l_{\infty}$ distance of 8/255 or a maximum $l_{2}$ distance of 0.05 as proposed in~\cite{athalye2018obfuscated}.

\subsection{Defense Requirements}
\label{sec:def-req}

We establish the following requirements for a qualified defense solution.
First, any modifications on the original DNN models are not allowed including retraining a model with different structures~\cite{papernot2016distillation} or datasets~\cite{tramer2017ensemble}. 
We set this requirement for two reasons. (1) Model retraining can significantly increase the computation cost, especially for large-scale DNN models (e.g. ImageNet~\cite{imagenet_cvpr09}). (2) Those defense methods lack generality to cover various types of attacks \cite{carlini2019evaluating}. 

Second, we consider adding preprocessing functions over the input samples before feeding them into the DNN models. Such operations can either remove the effects of adversarial perturbations on the prediction or make it infeasible for the adversary to generate AEs adaptively, even he knows every detail of the adopted mechanisms. The solution should be general-purpose and applicable to various types of data and DNN models of similar tasks. 

Third, the preprocessing functions should be lightweight with negligible computation cost to the inference pipeline. Besides, they should also preserve the usability of the original model without decreasing its prediction accuracy. Input preprocessing can introduce a trade-off between security and usability: the side effect of correcting the adversarial examples can also alter the prediction results of clean samples. 
A qualified solution should balance this trade-off with maximum impact on the adversarial samples and minimal impact on the clean ones.

\section{Data Augmentation Adopted in \name}
\label{sec:overview}
We introduce \name to mitigate existing adversarial attacks. \name leverages data augmentation techniques to preprocess the input samples and make them resistant against adversarial attacks. It consists of 15 data augmentation techniques classified into three categories. In this section, we detail the design and implementation of each method. Note that each method has its own parameters which can affect the defense results. We list the values of all these parameters following the method description, as well as summarized in~\tablename~\ref{tab: parameters}. The visual effect of each operation is demonstrated in \figurename~\ref{fig:alldefense}. We adopt the metrics of l2 norm, {Structural Similarity index}~(SSIM), and {Peak Signal-to-Noise Ratio}~(PSNR) in this figure to measure the amount of variations caused by these methods.



\subsection{Image Distortion}
\label{sec:imgdis}
This line of techniques conducts a distortion over pixels to remap each pixel away from their original positions. 
According to previous works \cite{athalye2018obfuscated, qiu2020mitigating}, a good image distortion approach can generate a large random variance between the original and preprocessed images with little impact on the model accuracy. This strategy can shatter the gradient and hence mitigate adversarial attacks in some ways. Moreover, image distortion techniques with randomness can break the basic assumption of the BPDA attack, which assumes the preprocessed image can be interpreted as an approximation of the original image~\cite{qiu2020mitigating}. We follow this strategy to present five novel image distortion functions in this category, as elaborated below. 


\subsubsection{Stochastic Affine Transformation (SAT)}
This operation adds randomness to the basic affine transformations. Three basic preprocessing procedures (translation, rotation, and scaling) are integrated to remove the adversarial perturbation on AEs. As a result, this method contains three coefficients for the three affine transformations: the translation limit $T$, the rotation limit $R$, and the scaling limit $S$. The coefficients used in \name are: $T=0.16$, $R=4$, and $S=0.16$.

For each image to be preprocessed, (1) it will be first shifted up by $\Delta_y$ pixels, and shifted left by $\Delta_x$ pixels. Here $\Delta_y$ and $\Delta_x$ are separately acquired by $\Delta = \delta \times l$, where $l$ is the side length and $\delta$ is acquired from a uniform distribution in the range of $(-T, T)$. (2) Then, the input sample will be rotated by $\delta_r \times \pi/180$ degrees, where $\delta_r$ is acquired from a uniform distribution in the range of $(-R, R)$. (3) Finally, the image's height and width will be scaled up to $\delta_s$ times, where $\delta_s$ is sampled from a uniform distribution in the range of $(1-S, 1+S)$. If $\delta_s$ is larger than 1, the final output will be obtained following a center cropping procedure to attain the original size. Otherwise, a center zero-padding will be adopted to output the original size. 


\subsubsection{Random Sized Cropping Affine (RSCA)}
Cropping is a widely adopted augmentation technique in the computer vision domain to inflate the training dataset \cite{szegedy2016rethinking, shorten2019survey}. \cite{guo2018countering} adopted random cropping with a kernel size of $90 \times 90$ to defeat gradient-based adversarial attacks 
In \name, we further develop \texttt{RSCA}, a cropping strategy with random sized kernels to enhance the defense results. \texttt{RSCA} has two coefficients: the minimum limitation $\theta$ and the aspect ration of cropping $\eta$. We choose $\theta=0.66$ and $\eta=0.91$ in \name to get the optimal defense results. 

This operation first sets the new height of the cropped image as $h_{new}\in (h\times \theta,h)$, and the new width as $w_{new}=\left \lfloor h_{new}\times \eta  \right \rfloor$. The coordinate $y_1$ for the crop is set as $y1=\left \lfloor (h-h_{new})\times {rand}_h \right \rfloor$, and $y_2$ is by adding $h_{new}$ to $y_1$. The coordinates $x_1$ and $x_2$ can be calculated following this same strategy with ${rand}_w$, where ${rand}_h$ and ${rand}_w$ are sampled independently from ${U}(0, 1)$. Then the original input is cropped with those four coordinates. Finally, the output image is obtained by resizing the image to its original size. 

\subsubsection{Random Sized Padding Affine (RSPA)}
Padding is another widely adopted data augmentation technique in deep learning tasks. In \cite{xie2018mitigating}, a random padding layer is added to the target model to protect the classification results from adversarial attacks. Following this strategy, we present an advanced random padding technique, \texttt{RSPA}. It has only one coefficient, the scaling limit $\lambda$. We set this value as 1.3 in our platform.

This operation first sets $\lambda_h = \lambda \times h$ as the maximum bound size for the scaling. Then it resizes the image to $(h_{new},h_{new} )$, where $h_{new}$ is an integer acquired from $U(h,\lambda_h)$. It randomly selects a pixel from the resized image as the center of padding and pad the image to the shape of $(\lambda_h,\lambda_h)$ with a value of 0.5. Finally, the output is acquired by resizing the image to the original shape. 

\subsubsection{Stochastic Elastic Transform (SET)}
\begin{figure*}[!htbp]
  \centering
  \includegraphics[width=\linewidth]{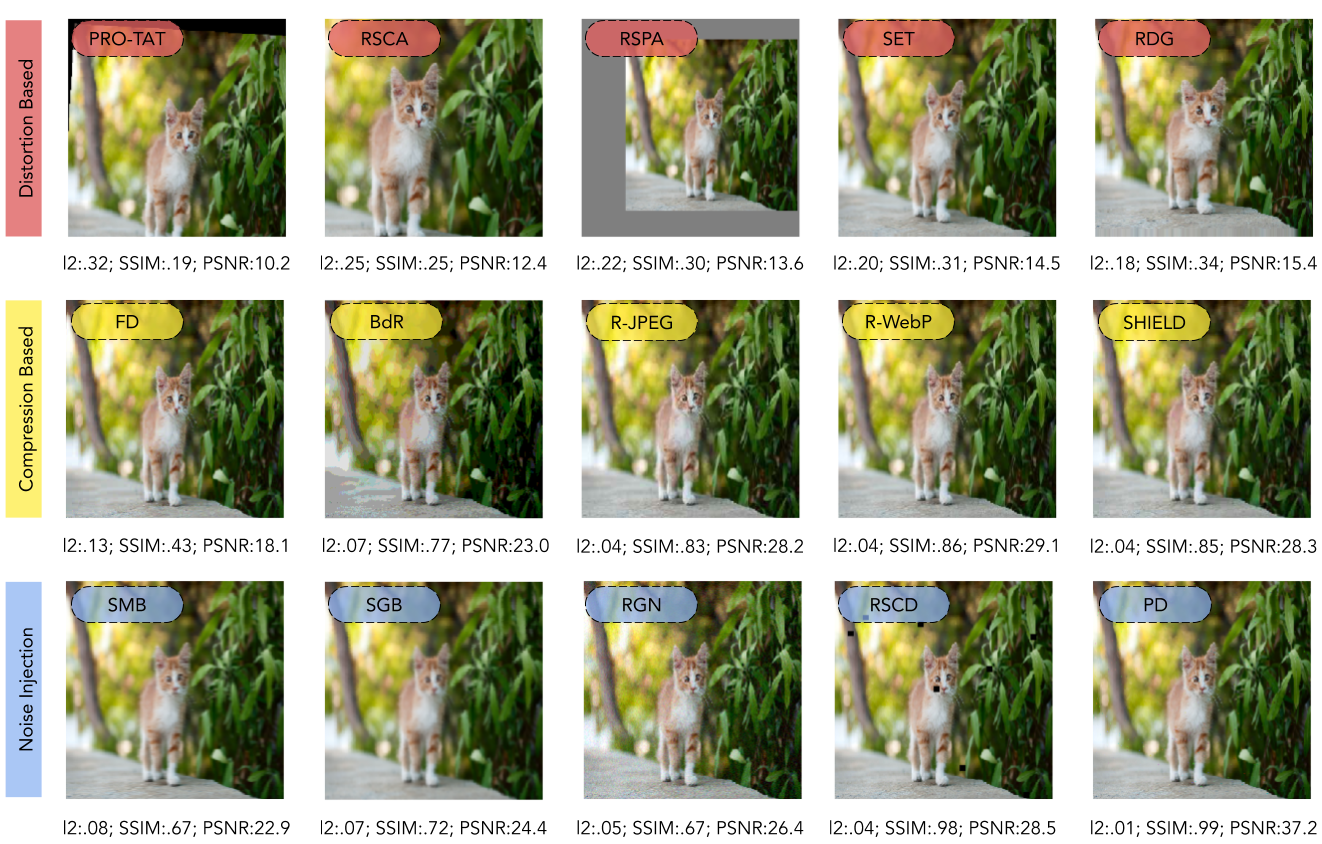}
  \caption{Visual results of 15 data augmentation techniques in three categories, with the corresponding L2, SSIM and PSNR values.}
  \label{fig:alldefense}
\end{figure*}

Simard et al. \cite{simard2003best}  proposed a data augmentation method, Elastic Distortions Transform, to expand the training dataset with respect to convolutional layers. With this operation, each pixel is displaced with elastic deformation, which can presumably remove adversarial perturbations. Elastic Transformation can randomly produce a more distorted image than the operations based on affine distortion (\texttt{SAT}, \texttt{RSCA}, \texttt{RSPA}, \texttt{RDG} \cite{qiu2020mitigating}). This can be visually observed in \figurename~\ref{fig:alldefense}. In \name, we design \texttt{SET}, an improved Elastic Transformation with randomness to further reduce the effects of adversarial attacks. \texttt{SET} can be interpreted as a two-stage process, where the first stage conducts a random basic affine transformation, followed by a random elastic distortion using Gaussian filter globally. Three coefficients are concerned in \texttt{SET}, the scale limit for the basic scaling procedure $\theta$, the standard deviation of the Gaussian kernel $\sigma$, and the elastic distortion scale index $\alpha$. The optimal values through testing are $\theta=20$, $\sigma=10$, and $\alpha=60$.

For each input, we first compute the expected affine effects by adding a random value $\delta \in U(-\theta, \theta)$ to the three pixels' coordinates. A random affine matrix is calculated based on the original coordinates and the processed coordinates. Then we apply this affine transformation globally according to the affine matrix. We generate two matrices of the same shape with each value randomly sampled from $U(-1,1)$. We compute the variation maps ($\delta_x$ and $\delta_y$) for horizontal and vertical coordinates, by multiplying $\alpha$ with the convolution result of the random matrices and Gaussian kernel with a standard deviation of $\sigma$. The maps for remapping is obtained by adding $\delta_x$ to each horizontal coordinates, and $\delta_y$ to each vertical coordinates. Using these new coordinates, \texttt{SET} finally outputs the distorted image.

\subsubsection{Random Distortion over Grids (RDG)}
This technique was first proposed in \cite{qiu2020mitigating} to invalidate the BPDA attack. 
The preprocessed images from \texttt{RDG} look quite similar to the ones from \texttt{SET}, as both of them conduct a remapping procedure to translate or distort the pixels away from each other. Different from \texttt{SET} which uses a Gaussian kernel to implement random distortion globall, \texttt{RDG} builds up grids vertically and horizontally to constrain the distortion in between. This can maintain a higher accuracy on clean samples. To tune RDG, two coefficients should be taken into consideration: the number of grids $d$, and the distortion limit $\delta$. The optimal coefficients used in \name are $d=26$, $\delta = 0.33$.

\texttt{RDG} first selects a corner to start generating $d$ grids uniformly. Between each two grids, it generates a distortion index $\Delta \in U(-\delta,\delta)$ and multiply it with the corresponding coordinates to get new coordinates. The newly coordinates is then used to map the original pixels to produce a distorted image. 

\subsection{Image Compression}
\label{sec:imgcom}

Most image compression techniques can incur non-differentiability to the inference phase due to the quantization effect \cite{qiu2020mitigating}. This can defeat advanced attacks against differentiable transformations, e.g., Expectation over Transformation \cite{athalye2018obfuscated}. In addition, lossy compression can drop certain amount of information from the original image, which can possibly remove the adversarial perturbation from AEs. We present five novel compression solutions for adversarial attack mitigation. Each method conducts a non-linear drop-value procedure to disturb the adversarial perturbation added on AEs. 


\subsubsection{Feature Distillation (FD)}
This method \cite{liu2019feature} is based on the standard JPEG compression. Since the adversarial perturbations could also exist inside the low-frequency bands which are normally ignored by naive JPEG, \texttt{FD} introduces an upgraded JPEG quantization operation to preprocess the input images. It measures the importance of input features for DNNs but not for human eyes by leveraging the statistical frequency component analysis within the DCT of JPEG. The adoption of this quantization operation can significantly improve the defense strength. This method can be further recursively deployed to improve the defense results.

\subsubsection{Bit-depth Reduction (BdR)}
This technique \cite{DBLP:conf/ndss/Xu0Q18} conducts a simple quantization procedure over the input image to squeeze the 8-bit (for each channel in the RGB image) images to fewer bits. For instance, for a sample from the ImageNet dataset, each pixel has three values for three channels of RGB ranging from 0 to 255, which can be represented by 8-bit. By conducting the shifting procedure to each pixel, we can attain images represented by fewer bits, thus the possible number of pixel values is reduced. Previous work \cite{DBLP:conf/ndss/Xu0Q18} demonstrated that \texttt{BdR} can achieve a satisfactory defense result against various standard adversarial attacks, especially for small-scale image datasets (MNIST and CIFAR-10). However, effectiveness of this method against more advanced attacks are never considered. In \name, we adopt the same coefficients as in \cite{DBLP:conf/ndss/Xu0Q18}, which downgrade the depth to 3-bit.

\subsubsection{SHIELD}
This approach \cite{das2018shield} aims to randomize the quantization operation by assigning a different quantization factor for each window during the JPEG compression process. The Stochastic Local Quantization (SLQ) method is used to divide an image into $8\times8$ blocks and apply a randomly selected JPEG compression quality (tuning quantization factors) to every block. \texttt{SHIELD} can achieve a slight improvement of classification accuracy than naive JPEG compression. However, it is unknown whether it can be used to defeat advanced adversarial attacks. In \name, we adopt the same coefficients as \cite{das2018shield}.

\subsubsection{Random JPEG Compression (R-JPEG)}
We designed \texttt{R-JPEG}, a novel compression technique by adding randomness to the quantization factor.
Different from \texttt{SHIELD} which adopts different quantization factors for different blocks, \texttt{R-JPEG} samples a random value from $U(Q_{min}, Q_{max})$ for all the windows as the quantization factor, where $Q_{min}$ and $Q_{max}$ are the two boundaries for the randomized quantization factor. As a result, the block generation and quantization value assignment is simplified, making \texttt{R-JPEG} more lightweight. Our evaluations indicate that \texttt{R-JPEG} can achieve a similar defense result as \texttt{SHIELD}. The coefficients are set as $Q_{min}=20$, $Q_{max}=80$ in \name.


\subsubsection{Random WebP Compression (R-WebP)}
This technique \cite{google2015webp} was originally proposed to improve the image compression rate, to meet the trend that high-resolution images are becoming more common in recent years \cite{toderici2015variable}. Lossy WebP compression adopts predictive coding to encode an image, which uses the values in neighboring blocks of pixels to predict the values in a block and then only encodes the difference \cite{google2015webp}. Similar to JPEG compression, WebP also includes a quantization procedure. Thus, a coefficient is also introduced to control the quantization scale. By adding randomness to this factor, \texttt{R-WebP} can achieve a better defense result against adversarial examples. Similar to \texttt{R-JPEG}, we sample a random quantization factor for each different input image from $U(Q_{min}, Q_{max})$.

\subsection{Image Noise Injection}
\label{sec:imgnoi}
This strategy normally randomizes the pixel values by injecting Gaussian noise to the image. We introduce five different techniques of noise injection, which can be used to defeat adversarial attacks. 


\subsubsection{Stochastic Motion Blur (SMB)}
Motion blur is ubiquitous in real-world photography, especially with resource-constrained devices, e.g., cell phones, on-board cameras \cite{gong2017motion}. Past works \cite{kurakin2016adversarial} have demonstrated that realistic settings like motion blur caused by certain angles and noise can invalidate adversarial examples. Hence, we propose \texttt{SMB}, which inject randomness to the process of a simulation of motion blur to remove adversarial perturbations. The motion blur in this method is achieved by convoluting a small-size normalized kernel globally.

Specifically, \texttt{SMB} first selects a random value $\phi$ from $U(3,\delta)$ as the size of the blurring kernel, where $\delta$ represents the blurring limit. Then a kernel with a random size of $(\phi,\phi)$ is generated by randomly drawing a straight line on the $(\phi,\phi)$ shaped white canvas. After normalizing the kernel, we move and convolute this kernel globally on the input to apply the motion blur. We set the value of $\delta$ as 9 to ensure model accuracy and defense effects concurrently.

\subsubsection{Stochastic Glass Blur (SGB)}
Glass Blur was proposed in \cite{hendrycks2019benchmarking} to evaluate the robustness of neural networks against common corruptions and perturbations. It simulates the effects of looking through a frosted glass. We add randomness into the Glass Blur process to further hinder the generation of adversarial examples. In \cite{hendrycks2019benchmarking}, only 5 sets of coefficients are adopted as the severity index ranging from 1 to 5. In \name, We further enlarge the set of coefficients to add higher randomness to the preprocessing. For each sample, Our Stochastic Glass Blur method conducts a stochastic initialization: it first selects a random standard deviation $\sigma$ from $U(0.7,1.5)$ as the Gaussian kernel, a random distance limit $\delta_{max}$ from from $[1,2,3,4]$ for pixels to be swapped , and a random number of interactions $I$ from $[1,2,3]$ to conduct the operation.

\subsubsection{Random Gaussian Noise (RGN)}

It has been shown that most of the Neural Networks are robust against certain amount of Gaussian Noise. Thus, the injection of Gaussian Noise to some extent can remove or obfuscate adversarial perturbations without harming the classification result. We improve this strategy by adding a constraint range with the standard deviation to ensure higher randomness and model accuracy at the same time. For each sample, our Random Gaussian Noise technique produces a random value from $\sigma_{min},\sigma_{max}$ as the standard deviation for the Gaussian Noise generator. Our evaluations suggest $\sigma_{min}=0.0005$, $\sigma_{max}=0.005$ and a mean value of $0$ for the Gaussian function as the optimal coefficients for the defense. 

\subsubsection{Random Sized Coarse Dropout (RSCD)}
Coarse Dropout and Cutout are simple image distortion methods to cut off small boxes from the original input. This method is evaluated for the robustness of a deep learning network in \cite{hendrycks2019benchmarking}. Since adversarial examples have less robustness than clean images \cite{guo2018countering}, this dropout strategy can potentially counter a number of adversarial attacks. We designed a new strategy, \texttt{RSCD}, which includes randomness to the coefficients of the Coarse Dropout procedure for better defense effects against adversarial attacks. \texttt{RSCD} randomly drops $n$ boxes, where $n$ is sampled from $\left \lfloor U(0,\lambda) \right \rfloor$. Each box is within the height of $h$ pixels and width of $w$ pixels, where $h$ and $w$ are sampled from $\left \lfloor U(1,\rho) \right \rfloor$ independently. The locations of those boxes over the entire image follow a uniform distribution ranging from $0$ to the original side length of the input. The optimal coefficients we evaluate to achieve the best defense effects are $\lambda=8$, and $\rho=8$.

\subsubsection{Pixel Deflection (PD)}
This technique \cite{prakash2018deflecting} adds random noise that are not sensitive to the DNN model to mitigate adversarial attacks. The conduction of deflection is to randomly sample a pixel from the input image and replace it with another randomly selected pixel within a small square neighborhood. Such a procedure can generate artificial noise that incur negligible impact on the DNN model, but disturb the adversarial perturbations. The introduction of such randomness can achieve satisfactory defense results against standard attacks \cite{prakash2018deflecting}. However, whether it can mitigate advanced attacks is never explored. We include it in \name as a potential (part of the) defense solution.



\section{Evaluations and Analysis}
\label{sec:evaluation}
In this section, we conduct comprehensive evaluations of all these data augmentation techniques in \name. Multiple kinds of adversarial attacks are considered: standard attacks (FGSM, IFGSM, C\&W, and LBFGS) and advanced attacks (PGD and BPDA). We present the configurations of our experiments, followed by the results and analysis. 

\subsection{Evaluation Setup}
\label{sec:setup}
We adopt Tensorflow 1.13.1 as the deep learning back-end framework to implement the preprocessing functions and the attacks. All the experiments were conducted on a server equipped with 8 Intel I7-7700k CPUs and 4 NVIDIA GeForce GTX 1080 Ti GPUs. 

\begin{table}
\centering
\caption{Parameters of the 15 methods in \name.}
\newcommand{\tabincell}[2]{\begin{tabular}{@{}#1@{}}#2\end{tabular}}
  \small
\begin{tabular}{c|c}
\Xhline{1pt}
\textbf{Defense} & \textbf{Parameter Values} \\

\Xhline{1pt}

\tabincell{c} {\texttt{SAT}} & $T=0.16$, $R=4$, $S=0.16$ \\
\hline 
\tabincell{c} {\texttt{RSCA}} & $\theta=0.66$, $\eta=0.91$ \\ \hline 
\tabincell{c} {\texttt{RSPA}} & $\lambda = 1.3$  \\ \hline 
\tabincell{c} {\texttt{SET}} & $\theta=20$, $\sigma=10$, $\alpha=60$ \\ \hline 
\tabincell{c} {\texttt{RDG} \cite{qiu2020mitigating}} & $d=26$, $\delta = 0.33$ \\

\Xhline{1pt}
\tabincell{c} {\texttt{FD} \cite{liu2019feature}} & - \\
\hline 
\tabincell{c} {\texttt{BdR} \cite{DBLP:conf/ndss/Xu0Q18}} & Bit depth = 3 \\
\hline 
\tabincell{c} {\texttt{R-JPEG}} & $Q_{min}=20$, $Q_{max}=80$\\
\hline 
\tabincell{c} {\texttt{R-WebP}} & $Q_{min}=20$, $Q_{max}=80$\\
\hline 

\tabincell{c} {\texttt{SHIELD} \cite{das2018shield}} & - \\

\Xhline{1pt}
\tabincell{c} {\texttt{SMB}} & $\delta = 9$\\
\hline 
\tabincell{c} {\texttt{SGB}} & $\delta_{max} \in [1,2,3,4]$, $I \in [1,2,3]$\\
\hline 
\tabincell{c} {\texttt{RGN}} & $\sigma_{min}=0.0005$, $\sigma_{max}=0.005$ \\
\hline 
\tabincell{c} {\texttt{RSCD}} & $\lambda=8$, $\rho=8$\\
\hline 
\tabincell{c} {{\texttt{PD}}\cite{prakash2018deflecting}} & - \\

\Xhline{1pt}
\end{tabular}
\label{tab: parameters}
\end{table}

\noindent\textbf{Model \& dataset.}
The techniques in \name are designed to be general-purpose and can be applied to various models for computer vision tasks. Without the loss of generality, we consider a pre-trained Inception V3 model \cite{szegedy2016rethinking} over the ImageNet dataset as the target model of attacks and defenses. This state-of-the-art model can reach 78.0\% top-1 and 93.9\% top-5 accuracy. We randomly select 100 images from the ImageNet Validation dataset for AE generation. These images can be predicted correctly by this Inception V3 model. 

\begin{table*}
\centering
\caption{ACC of different augmentation techniques against different attacks.}
\newcommand{\tabincell}[2]{\begin{tabular}{@{}#1@{}}#2\end{tabular}}
\begin{tabular}{c|c|c|c|c|c|c|c|c}
\Xhline{1pt}
\textbf{Defense} & \textbf{Clean} & \tabincell{c}{\textbf{I-FGSM}\\($\epsilon$=.03)} & \textbf{C\&W} & \textbf{LBFGS} & \tabincell{c}{\textbf{PGD}\\($l_2$=.05)} &\tabincell{c}{\textbf{BPDA}\\($R$=10)} & \tabincell{c}{\textbf{BPDA}\\($R$=50)} & \tabincell{c}{\textbf{BPDA}\\($l_2$=.05)}\\
\Xhline{1pt}
\tabincell{c} {{Baseline}} & {1.00}  & {0.02} & {0.00} & {0.00} & {0.00} & {0.06} & {0.00} & {0.00}\\
\Xhline{1pt}

\tabincell{c} {\texttt{SAT}} & \textbf{0.98}  & 0.85 & 0.78& \textbf{0.96} & 0.66 & \textbf{0.90}&\textbf{0.82} &\textbf{0.73}\\
\hline 
\tabincell{c} {\texttt{RSCA}} & \textbf{0.98}  & 0.85 & 0.78 & \textbf{0.96} & 0.72 & \textbf{0.90}& \textbf{0.82} &\textbf{0.75}\\
\hline 
\tabincell{c} {\texttt{RSPA}} & 0.96  & \textbf{0.87} & \textbf{0.85} &\textbf{0.96} & 0.76& 0.80& 0.67 &0.60\\
\hline 
\tabincell{c} {\texttt{SET}} & 0.95  & 0.84 & \textbf{0.84} & 0.94 & 0.69& 0.85& 0.71&0.61\\
\hline 
\tabincell{c} {\texttt{RDG} \cite{qiu2020mitigating}} & 0.95  & \textbf{0.87} & 0.78 & 0.95 & 0.73&\textbf{0.86} & \textbf{0.76} &0.52\\

\Xhline{1pt}
\tabincell{c} {\texttt{FD} \cite{liu2019feature}} & \textbf{0.97}  & 0.86 & 0.80 & \textbf{0.97} & \textbf{0.77}&0.81 &0.54 & 0.39\\
\hline 
\tabincell{c} {\texttt{BdR} \cite{DBLP:conf/ndss/Xu0Q18}} & 0.92  & 0.82 & 0.61 & 0.90 & 0.62&0.13 & 0.00&0.00\\
\hline 
\tabincell{c} {\texttt{R-JPEG}} & 0.93 & 0.85 & 0.77 & 0.93 & 0.76&0.44 & 0.28&0.26\\
\hline 
\tabincell{c} {\texttt{R-WebP}} & 0.94 & 0.85 & 0.79 & 0.94 & \textbf{0.78}& 0.46& 0.28&0.21\\
\hline 

\tabincell{c} {\texttt{SHIELD} \cite{das2018shield}} & 0.93 & \textbf{0.87} & 0.81 & 0.93 & 0.76& 0.44& 0.15&0.14\\

\Xhline{1pt}
\tabincell{c} {\texttt{SMB}} & 0.92 & 0.78 & 0.71 & 0.87 & 0.63& 0.51& 0.35&0.30\\
\hline 
\tabincell{c} {\texttt{SGB}} & 0.83 & 0.75 & 0.70 & 0.81 & 0.70& 0.12& 0.06&0.04\\
\hline 
\tabincell{c} {\texttt{RGN}} & 0.94 & \textbf{0.89} & 0.64 & 0.94 & 0.76& 0.47& 0.31&0.26\\
\hline 
\tabincell{c} {\texttt{RSCD}} & 0.94 & 0.38 & 0.15 & 0.77 & 0.01& 0.19& 0.10&0.06\\
\hline 
\tabincell{c} {{\texttt{PD}}\cite{prakash2018deflecting}} & 0.96 & 0.30 & 0.11 & 0.89 & 0.00& 0.13& 0.04&0.03\\

\Xhline{1pt}
\end{tabular}
\label{tab: ACC}
\end{table*}

\begin{table*}[!htbp]
\centering
\caption{ASR of different augmentation techniques against different attacks.}
\newcommand{\tabincell}[2]{\begin{tabular}{@{}#1@{}}#2\end{tabular}}

\begin{tabular}{c|c|c|c|c|c|c|c|c}
\Xhline{1pt}
\textbf{Defense} & \textbf{Clean}  & \tabincell{c}{\textbf{I-FGSM}\\($\epsilon$=.03)} & \textbf{C\&W} & \textbf{LBFGS} & \tabincell{c}{\textbf{PGD}\\($l_2$=.05)} &\tabincell{c}{\textbf{BPDA}\\($R$=10)} & \tabincell{c}{\textbf{BPDA}\\($R$=50)} & \tabincell{c}{\textbf{BPDA}\\($l_2$=.05)}\\
\Xhline{1pt}
\tabincell{c} {{Baseline}} & {0.00}  & {0.95}& {1.00}&{1.00}& {1.00}&{0.87}& {1.00} &{1.00}\\
\Xhline{1pt}

\tabincell{c} {\texttt{SAT}} & 0.00  & 0.02 & 0.01& 0.00 & 0.05& 0.00 & 0.00 &0.00\\
\hline 
\tabincell{c} {\texttt{RSCA}} & 0.00  & 0.01 & 0.01 & 0.00 & 0.04&0.00& 0.00&0.00\\
\hline 
\tabincell{c} {\texttt{RSPA}} & 0.00  & 0.02 & 0.02 & 0.00 &0.02& 0.01& 0.04&0.04\\
\hline 
\tabincell{c} {\texttt{SET}} & 0.00 & 0.01 & 0.01 & 0.00 &0.01& 0.00& 0.00&0.00\\
\hline 
\tabincell{c} {\texttt{RDG} \cite{qiu2020mitigating}} & 0.00  & 0.00 & 0.00 & 0.00 &0.01&0.00 & 0.01&0.01\\

\Xhline{1pt}
\tabincell{c} {\texttt{FD} \cite{liu2019feature}}  & 0.00 & 0.00 & 0.06 & 0.00 &0.04&0.02 &0.16 &0.27\\
\hline 
\tabincell{c} {\texttt{BdR} \cite{DBLP:conf/ndss/Xu0Q18}}  & 0.00 & 0.02 & 0.20 & 0.00 &0.07&0.75 & 1.00&1.00 \\
\hline 
\tabincell{c} {\texttt{R-JPEG}}  & 0.00 & 0.00 & 0.03 & 0.00 &0.00&0.23 & 0.44&0.47 \\
\hline 
\tabincell{c} {\texttt{R-WebP}}  & 0.00 & 0.00 & 0.01 & 0.03 & 0.02&0.22& 0.45&0.48 \\
\hline 

\tabincell{c} {\texttt{SHIELD} \cite{das2018shield}}  & 0.00 & 0.01 & 0.02 & 0.00 & 0.01&0.23& 0.63&0.65 \\

\Xhline{1pt}
\tabincell{c} {\texttt{SMB}}  & 0.00 & 0.08 & 0.01 & 0.00 & 0.12&0.14& 0.23&0.29 \\
\hline 
\tabincell{c} {\texttt{SGB}} & 0.00 & 0.01 & 0.00 & 0.00 & 0.01&0.67& 0.78&0.84 \\
\hline 
\tabincell{c} {\texttt{RGN}}  & 0.00 & 0.00 & 0.20 & 0.00 & 0.05&0.14& 0.39&0.39 \\
\hline 
\tabincell{c} {\texttt{RSCD}}  & 0.00 & 0.45 & 0.78 & 0.05 & 0.96&0.48& 0.62&0.70 \\
\hline 
\tabincell{c} {{\texttt{PD}}\cite{prakash2018deflecting}}  & 0.00 & 0.60 & 0.89 & 0.02 & 1.00&0.70& 0.90&0.91 \\
\Xhline{1pt}
\end{tabular}
\label{tab: ASR}
\end{table*}

\noindent\textbf{Attacks.}
All adversarial attacks are conducted in a targeted fashion, where the targeted labels are randomly selected different from the original ones. We consider an attack is successful only if the prediction of the model is the targeted class. Cleverhans 2.1.0 is utilized to conduct standard adversarial attacks (I-FGSM, C\&W, and LBFGS) over the target model\footnote{We did not report the attack results of FGSM as it is a relatively weaker approach~\cite{carlini2019evaluating}, and the results are similar as I-FGS}. For I-FGSM, AEs are generated under $l_\infty$ constraint with $\epsilon= 0.03$. I-FGSM is iterated ten times. For LBFGS and C\&W, the optimization process is iterated until all desired AEs are discovered under the $l_2$ constraint; namely $l_2<0.05$. For LBFGS, the binary search step is set as 5, and the maximum number of iterations is set as 1000. For C\&W, the binary search step is set as 5, the maximum number of iterations is set as 1000, and the learning rate is 0.1. For PGD and BPDA, the learning rate is set as 0.1. For PGD, we continue the generation process utill the $l_2$ of the AE reaches the constraint $0.05$. For BPDA, we adopt an early stop process: the attack stops when an qualified AE (success) or it hits the $l_2$ bound (failure).

\noindent\textbf{Defenses.}
The implementations of \texttt{R-JPEG}, \texttt{R-WebP}, \texttt{SMB}, \texttt{SGB}, \texttt{RGN}, and \texttt{RSCD} are based on the library of Albumentations 0.4.5. The values of all the coefficients are listed in Table \ref{tab: parameters}. For the standard attacks (FGSM, IFGSM, C\&W, and LBFGS) and the PGD attack, the selected preprocessing function is only applied once to the generated adversarial examples during inference. For the BPDA attack, we apply the preprocessing for each round of gradient calculation. 

\noindent\textbf{Metrics.}
The pixel values of the image are normalized to $[0, 1]$. We use the $l_{2}$-norm and $l_{inf}$-norm to measure the scale of perturbations generated by each attack. The $l_2$-norm is calculated by computing the total root-mean-square distortion normalized by the number of pixels ($299 \times 299 \times 3$). The $l_{inf}$-norm is calculated as the maximum change of all the pixels in one image.
We only accept adversarial examples with a $l_2$ norm smaller than 0.05 (C\&W, LBFGS, PGD, and BPDA) or within the $l_{inf}$ bound of 0.03 (I-FGSM).
For each attack, we measure the prediction accuracy of the generated AEs (ACC) and the attack success rate (ASR). 
A higher ACC or smaller ASR reflects higher robustness and effectiveness of the preprocessing function. 

\subsection{Evaluation Results}
Tables \ref{tab: ACC} and \ref{tab: ASR} show the ACC and ASR of different augmentation techniques under different attacks. The first row of each table shows the baseline case without any defense methods. We can observe dramatic a drop in the ACC and increase in ASR for all the adversarial attacks. Most attacks can achieve close to 0\% ACC and 100\% ASR. 


The first column (``Clean'') of each table denotes the inference results over the benign samples. Most preprocessing functions can maintain high ACCs on clean samples, making them functionality-preserving. Also, their ASRs all stay at 0, indicating that no operations will incur false positives. In a nutshell, these results show that the 15 augmentation methods in \name have slight influence on the clean samples. 

The rest rows in the two tables denote the attack results for the preprocessing techniques in \name. 
The bold data in each column in Table \ref{tab: ACC} denote the top-2 ACCs of the preprocessing functions. 
(1) Generally we observe that image distortion operations have the best robustness. Their ASRs are close to zero under 100 rounds of BPDA, which is regarded as a very strong attack. (2) The image compression operations have comparable effectiveness. They demonstrate the best performance among these categories for PGD attack. However, they are vulnerable to BPDA attack even under 10 rounds.
\texttt{FD} still performs better than other image compression techniques in defeating BPDA as it has a fine-tuned quantization matrix to deflect adversarial perturbation.
(3) The noise injection solutions achieves similar results over I-FGSm attacks as the first category. They have relatively worse performance on both ACC and ASR for the other attacks. 

Particularly, we are interested in BPDA, one of the strongest up-to-date attacks. We conducted a $l_2$ bounded attack. We observe that only the image distortion techniques can maintain acceptable ACCs. Their ASRs are also constrained to close to zero. 
Besides these five techniques, only \texttt{FD} and \texttt{SMB} can incur ACCs higher than 30\%, but the ASRs are also higher than 27\%, making them less effective.


\subsection{Discussion}

Based on the above results, if we just select one preprocessing function as our defense solution, then the image distortion strategy is recommended. It achieves the best defense performance to maintain high ACC and low ASR than the other two. Specifically, for the advanced approximation gradient-based attack, the ACCs of the image compression and noise injection operations are much lower. The main reason is that the image distortion operations can generate high-level randomness when preprocessing the input images. For instance, \texttt{RDG} can not only randomly change the pixel values but also alter the pixel positions and drop a certain ratio (more than 20\%) of pixels. 
Such randomness will bring a shattered gradient for inference session, and make the gradient approximation more difficult. Therefore, deploying the BPDA method in~\cite{athalye2018obfuscated} cannot generate effective AEs even when the $l_{2}$ norm reaches the upper bound of 0.05. 

However, this random image distortion can be threatened by another powerful attack, i.e., EOT~\cite{athalye2018obfuscated}, which calculates the average values of gradients to eliminate the random effects. Even though this attack requires the preprocessing function to be differentiable, a non-differentiable random data augmentation method is still vulnerable to the concatenation of BPDA and EOT attack. 
For instance, \texttt{RDG} can be broken by the EOT attack with ACC lower than 20\%. 

To overcome the above challenges, a more sophisticated strategy is to combine multiple data augmentation techniques. In Section~\ref{sec:case}, we demonstrate a case study about how to select certain functions from \name to mitigate the BPDA+EOT attack


\section{Case Study: Ensemble of Augmentations}
\label{sec:case}

As discussed in Section~\ref{sec:evaluation}, it is difficult to employ just one data augmentation technique to defeat all adversarial attacks. Inspired~\cite{raff2019barrage}, we can build a combination of multiple methods to enhance the defense strength. In this section, we present a case study, to show how to select appropriate augmentation functions from \name to establish a more effective solution against advanced attacks, including BPDA, EOT, BPDA+EOT and adaptive attacks.


\subsection{Methodology}
The key insight is to use a combination of multi-layer data augmentation techniques to defeat adversarial attacks. The challenge is how to select the optimal functions to achieve this goal. Since each augmentation function is effective against standard attacks, we analyze the mechanisms of advanced attacks to form the desired solution.


As mentioned in Section~\ref{sec:bg}, the main idea of the BPDA attack is to replace a non-differentiable preprocessing function $g(\cdot)$ with a differentiable one to approximate the gradients. The common practice is to use the identify function to approximate the gradients during back propagation $g(x)\approx x$. The whole process is described in Eq. \ref{eq:bpda1}. 

\begin{subequations}
\label{eq:bpda1}
  \begin{align}
\bigtriangledown _{x}f(g(x))| _{x = \hat{x}} & = \bigtriangledown _{x} f(x)| _{x = g(\hat{x})} \times \bigtriangledown _{x} g(x) | _{x = \hat{x}} \\
& \approx \bigtriangledown _{x} f(x)| _{x = g(\hat{x})} \times \bigtriangledown _{x} x | _{x = \hat{x}} \\
& = \bigtriangledown _{x} f(x)| _{x = g(\hat{x})}
  \end{align}
\end{subequations}

Another advanced attack, EOT, is able to handle the random transformations by statistically computing the gradients over the expected transformation to the input $x$. This is able to defeat the random image distortion solutions. 
Formally, for a preprocessing function $g(\cdot)$ that randomly transforms $x$ from a distribution of transformations $T$, EOT optimizes the expectation over the transformation with respect to the input by $\mathbb{E} _{t\sim T}f(g(x))$, as shown in Eq. \ref{eq:eot1}. The adversary can then get a proper expectation with samples at each gradient descent step.

\begin{equation}\label{eq:eot1}
    \bigtriangledown_{x}  \mathbb{E} _{t\sim T}f(g(x)) = \mathbb{E} _{t\sim T} \bigtriangledown_{x} f(g(x))
\end{equation}


Based on the above analysis of two attacks, we can build the defense ensemble as below: we select two data augmentation functions from \name and concatenate them as our solution: $g(\cdot) = g_1 \circ g_2$. Function $g_1$ is a highly randomized function that can mitigate the BPDA attack. This can be selected from the image distortion category. Function $g_2$ is a non-differentiable function to thwart the EOT attack. This can be selected from the image compression category. Such a combination can significantly enhance the difficulty of gradient calculation and AE generation. Below we demonstrate the evaluations to validate the effectiveness of our strategy. 


\subsection{Evaluation}
\label{sec:casestudyEva}

\subsubsection{Advanced Attacks}
We follow the same setup in Section~\ref{sec:setup}. We consider the EOT and BPDA+EOT attacks. For EOT, we select an ensemble size of 30 for calculating the expected gradients\footnote{We tested different ensemble sizes from 2 to 40. These sizes have little influence on ASR and ACC. A larger ensemble size can generate AEs smaller than $l_{2}$.}. We select \texttt{RDG} as $g_1$ and \texttt{FD} as $g_2$. We measure the performance of individual \texttt{RDG}, \texttt{FD} and their combination \texttt{FD+RDG} under BPDA, EOT and BPDA+EOT attacks. The ACC and ASR are shown in~\tablename~\ref{tab:accFDRDG} and~\tablename~\ref{tab:asrFDRDG}, respectively.


We observe that a single \texttt{RDG} is vulnerable to EOT with an ACC of 0.09 and ASR of 0.82. 
\texttt{FD} is able to defeat EOT, but has a low ACC (0.39) for the BPDA bounded attack. 
However, if we concatenate the \texttt{RDG} and \texttt{FD} together as \texttt{FD+RDG}, we can get a good defense solution that significantly outperforms each individual function: it has a better ACC than  \texttt{RDG} under the BPDA bounded attack and can maintain ACC as 0.60 for 50 rounds of the BPDA+EOT attack. Moreover, the ASR of BPDA+EOT at 50 rounds is only 0.06. 
We perform more rounds of the attack until the images with adversarial perturbations reach the bound ($l_2 = 0.05$). \texttt{FD+RDG} is able to maintain a model accuracy of 58\% and reduce the attack success rate below 7\%. 

\begin{table}[!htbp]
  \centering
  \newcommand{\tabincell}[2]{\begin{tabular}{@{}#1@{}}#2\end{tabular}}
  \caption{ACC of the proposed multi-layer data augmentation \texttt{FD+RDG}.}
    \begin{tabular}{c|c|c|c|c}
    \toprule
    \tabincell{c}{Method} & \tabincell{c}{\textbf{BPDA}\\($R$=50)}& \tabincell{c}{\textbf{BPDA}\\($l_2$=.05)}&
    \tabincell{c}{\textbf{EOT} }&
    \tabincell{c}{\textbf{BPDA+EOT}\\($R$=50)}\\
    \midrule
    \texttt{RDG} & 0.76 & 0.52 & 0.09 & -  \\
    \texttt{FD} & 0.54 & 0.39 & Nan & -  \\
    \texttt{FD+RDG} & 0.71 & 0.62 & - & 0.60 \\
    \bottomrule
    \end{tabular}
  \label{tab:accFDRDG}
\end{table} 

\begin{table}[!htbp]
  \centering
  \newcommand{\tabincell}[2]{\begin{tabular}{@{}#1@{}}#2\end{tabular}}
  \caption{ASR of the proposed multi-layer data augmentation \texttt{FD+RDG}.}
    \begin{tabular}{c|c|c|c|c}
    \toprule
    \tabincell{c}{Method} & \tabincell{c}{\textbf{BPDA}\\($R$=50)}& \tabincell{c}{\textbf{BPDA}\\($l_2$=.05)}&
    \tabincell{c}{\textbf{EOT}}&
    \tabincell{c}{\textbf{BPDA+EOT}\\($R$=50)}\\
    \midrule
    \texttt{RDG} & 0.01 & 0.01 & 0.82 & -  \\
    \texttt{FD} & 0.16 & 0.27 & Nan & -  \\
    \texttt{FD+RDG} & 0.00 & 0.02 & - & 0.06 \\
    \bottomrule
    \end{tabular}
  \label{tab:asrFDRDG}
\end{table}

\begin{figure}[!htbp]
  \centering
  \includegraphics[width=0.95\linewidth]{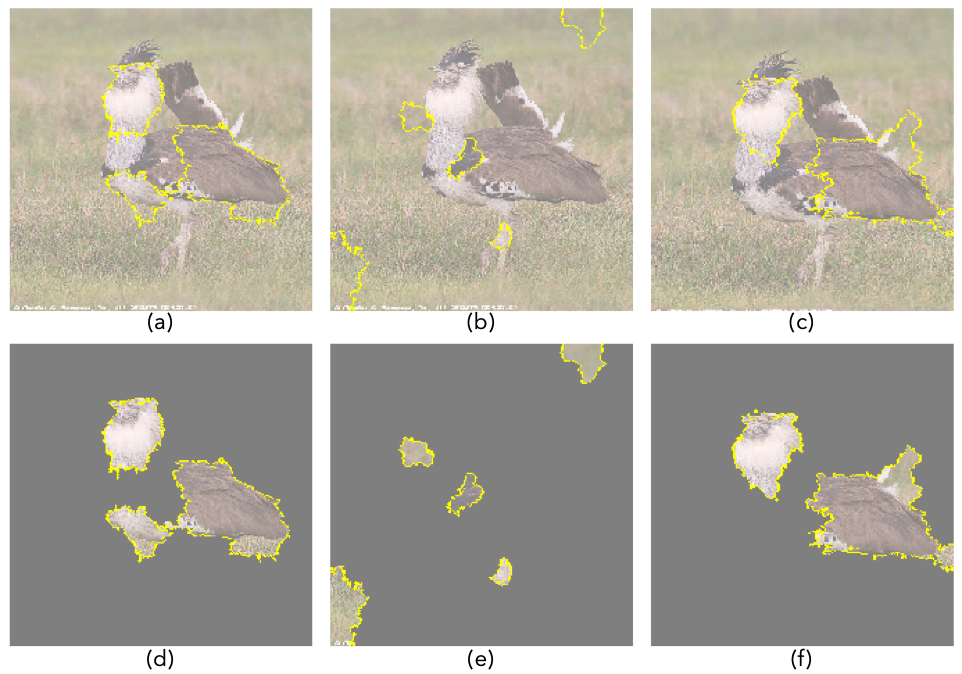}
  \caption{Explaining the BPDA+EOT adversarial example and preprocessing function using LIME.}
  \label{fig:lime}
\end{figure}

\noindent\textbf{Visual explanations.}
We use Local Interpretable Model-Agnostic Explanations \cite{ribeiro2016why} to demonstrate the mechanisms and effects of our solution. Figure \ref{fig:lime} shows the interpretation results of a normal sample (a), the corresponding AE (b) generated from BPDA+EOT, and the preprocessed output from \texttt{FD+RDG} (c). The highlighted parts denote the critical regions that determine the classification results. Figures \ref{fig:lime} (d) - (f) show the corresponding regions only for better comparisons. We observe the explanation regions of the AE are quite distinct from the original image. After preprocessing with \texttt{FD+RDG}, the explanation regions are similar as the clean image again. This shows the preprocessing function can indeed revert the malicious perturbations to normal. 

\subsubsection{More Sophisticated Adaptive Attacks}

\begin{figure}[!htbp]
\centering
\subfigure[ACC per round.] 
{\includegraphics[width = 0.49\linewidth]{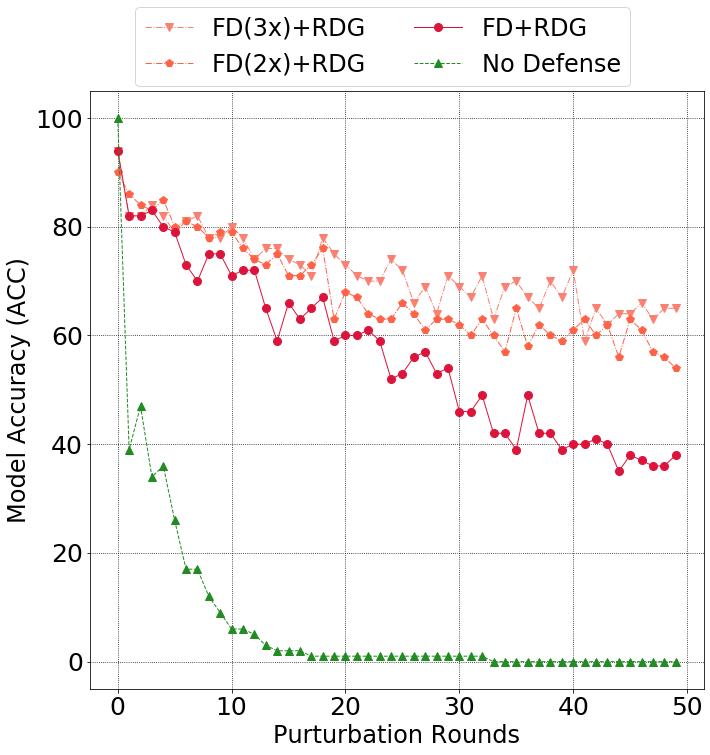}\label{newEOT_acc}}
\subfigure[ASR per round.] 
{\includegraphics[width = 0.49\linewidth]{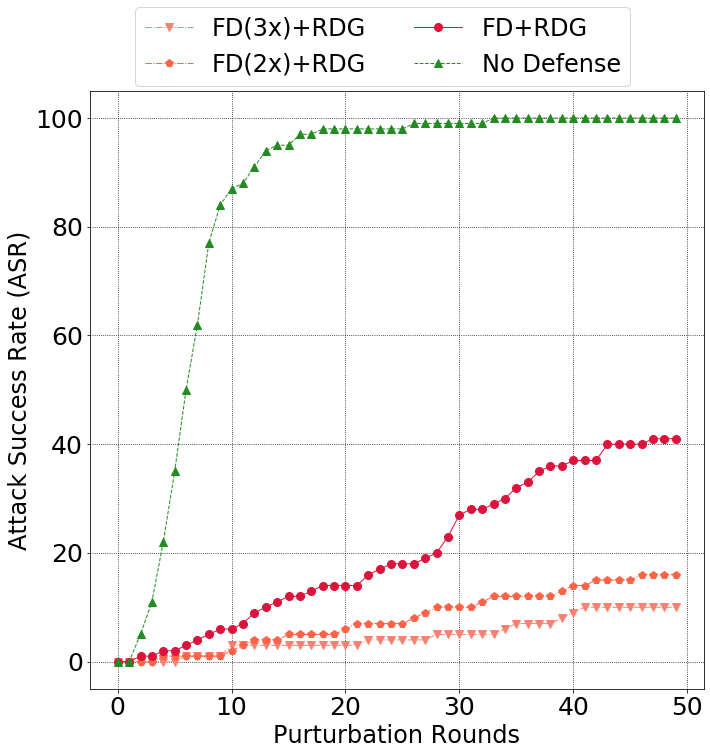}\label{newEOT_suc}}
\caption{Performance of the adaptive attack based on EOT.}
\label{Evaluation}
\end{figure}

We consider more sophisticated attacks. Although the adversary cannot use EOT to attack the entire preprocessing function $g(\cdot)$, it is possible that he can generate AEs only based on the differentiable component. In our case, the adversary can ignore the existence of $g_2$ and perform EOT on the operation $g_1$ only. With the generated AE, the adversary can test if it can defeat $g_2$ as well. 
He can repeat the above procedure until a proper AE is found that can fool both $g_1$ and $g_2$. \figurename~\ref{newEOT_acc} and \ref{newEOT_suc} show the ACC and ASR of this adaptive attack (red line with dots) versus the baseline case without any defenses (green line with triangle). 
We can observe that after 50 rounds, the ACC of \texttt{FD+RDG} is 40\% and the ASR is 40\%. 

Although this attack can defeat \texttt{FD+RDG} to some extent, it is not very practical as the cost of generating such adversarial examples is very high. It is worth noting that in this attack,  
EOT is used to target the randomization operation $g_1$ only. For the non-differentiable operation $g_2$ (i.e., \texttt{FD}), the adversary has to use a brute-force fashion to handle: trying different AEs generated from EOT targeting only $g_1$ until an AE is discovered that can fool $g_1 \circ g_2$. 
Its attack success rate is highly dependent on the robustness of $g_2$: a stronger $g_2$ can significantly increase the attack cost or even totally mitigate it. 
For instance, we can follow the idea in \cite{liu2019feature} to stack one extra \texttt{FD} operation in $g$: \texttt{FD$\times 2$+RDG}. Then the ACC after 50 rounds increases to 55\% and the ASR drops to 17\% (orange line with pentagon in Figure \ref{Evaluation}). 
We can further increase the robustness of $g_2$ with \texttt{FD$\times 3$} operations, and the defense results are better (orange line with triangle). 
\tablename~\ref{tab:asr_round} shows the attack rounds required to reach specific ASRs. This can also validate our conclusions. 

\begin{table}[!htbp]
  \centering
  \caption{Adaptive attacks against different defenses.}
    \begin{tabular}{cccccc}
    \toprule
    \multirow{2}[2]{*}{Method} & \multicolumn{5}{c}{Attak rounds required to target ASR} \\
    \cline{2-6}
          & 10\%  & 30\%  & 50\%  & 70\%  & 90\% \\
    \midrule
    
    \texttt{FD+RDG} & 14 & 38 & 59 & 207 & 7424 \\
    \texttt{FD$\times 2$+RDG} & 31 & 83 & 538 & 10000+ & 10000+ \\
    \texttt{FD$\times 3$+RDG} & 57 & 216 & 10000+ & 10000+ & 10000+ \\
    \bottomrule
    \end{tabular}
  \label{tab:asr_round}
\end{table}

\subsection{Discussion}

We have shown that utilizing one single data augmentation function is hardly enough to provide sufficient security protection against various adversarial attacks. Instead, we can collect multiple functions from \name based on the attack characteristics to build a more robust solution. 
Here, we do not claim that \texttt{FD+RDG} or \texttt{FD$\times 3$+RDG} is the ultimate solution for all existing adversarial attacks. Instead, we use this study to illustrate that (1) combination of less secure augmentation functions can possibly give a stronger defense~\cite{raff2019barrage}; (2) design of more sophisticated defenses should consider the fundamental mechanisms of the target attacks. With these two points, we expect researchers can utilize \name to design more effective solutions against new attacks in the future. 


\section{Conclusion}
\label{sec:conclusion}

In this paper, we design \name, a comprehensive platform for defenses of adversarial examples. It consists of 15 data augmentation functions from three different categories. Extensive evaluations show that these functions can be used to defeat various kinds of adversarial attacks. We also provide a case study to show that combination of certain functions can yield more effective solutions to mitigate more advanced attacks. We believe \name can be a valuable platform for researchers and practitioners to understand the mechanisms of adversarial attacks and defenses, and to build more efficient and effective defenses for robustness enhancement of DNN models.


\bibliographystyle{IEEEtran}
\bibliography{mybibfile}

\end{document}